\documentclass{article}
\usepackage[bottom]{footmisc}
\usepackage[disable]{todonotes}
\usepackage{graphicx}
\usepackage{threeparttable}
\usepackage{amsmath,amsfonts,amssymb,float}
\usepackage{graphicx}
\graphicspath{{./images/}}
\usepackage{subcaption}
\usepackage{caption}
\usepackage{comment}
\usepackage{placeins}
\usepackage{algorithm2e}
\usepackage{authblk}
\usepackage{blindtext}
\usepackage{hyperref}
\usepackage{etoolbox}
\usepackage{adjustbox}

\begin{document}
\title{Generative Models For Deep Learning with Very Scarce Data}

\author[1]{ Juan Maro\~nas jmaronas@prhlt.upv.es\thanks{correspondence to: Juan Maro\~nas $<$jmaronas@prhlt.upv.es$>$,$<$jmaronasm@gmail.com$>$. Work published at 23th Iberoamerican Congress on Pattern Recognition}}
\author[1]{Roberto Paredes rparedes@prhlt.upv.es}
\author[2]{Daniel Ramos daniel.ramos@uam.es}
\affil[1]{Pattern Recognition and Human Language Technology, Universitat Polit\`ecnica de Val\`encia, Valencia , Spain}
\affil[2]{AUDIAS , Universidad Aut\'onoma de Madrid, Madrid, Spain}
\date{November 2018}
\providecommand{\keywords}[1]{\textbf{Keywords:} #1}
\maketitle

\begin{abstract}
  The goal of this paper is to deal with a data scarcity scenario
  where deep learning techniques use to fail. We compare the use of
  two well established techniques, Restricted Boltzmann Machines and
  Variational Auto-encoders, as generative models in order to increase
  the training set in a classification framework. Essentially, we rely on Markov Chain
  Monte Carlo (MCMC) algorithms for generating new samples. We show
  that generalization can be improved comparing this methodology to other
  state-of-the-art techniques, e.g. semi-supervised learning with
  ladder networks. Furthermore, we show that RBM is better than VAE generating new
  samples for training a classifier with good generalization capabilities.\\
  
\end{abstract}

\section{Introduction}

In the last few years deep neural networks have achieved
state-of-the-art performance in many task such as image recognition
\cite{DBLP:journals/corr/SzegedyIV16}, object recognition
\cite{1506.02640}, language modeling
\cite{DBLP:journals/corr/abs-1301-3781}, machine translation
\cite{DBLP:journals/corr/SutskeverVL14} or speech recognition
\cite{hinton16speechprocessing}. One of the key facts that increased
this performance is the great amount of available data. This amount of
data together with the high expressiveness of neural networks as functions
approximators and appropriate hardware lead us to an
unprecedented performance in challenging problems.
\newline

However, deep learning lacks of success in scenarios where the amount
of labeled data is scarce. In this work we aim at providing a
methodology in order to apply deep learning techniques to problems with
\emph{very} scarce available data. Some techniques are proposed to deal
with such data size problem: semi supervised learning
techniques such as the ladder network \cite{NIPS2015_5947}, Bayesian
modeling \cite{1506.02158} and data augmentation (DA)
\cite{NIPS2017_6872}. In particular, data augmentation uses to be referred to the
techniques where the practitioners know the most common data
variability, as in image recognition, and these variations can be
applied to the available data in order to obtain new samples. On the
other hand, there are other methods not assisted by practitioners to
generate new samples: generative adversarial networks, GANs
\cite{1406.2661}, variational models such as variational auto-encoder
VAE \cite{1401.4082,1312.6114} and autoregressive models
\cite{1606.05328}.
\newline

\begin{figure}[!b]
\begin{subfigure}{0.5\textwidth}
\centering
  \includegraphics[scale=0.3]{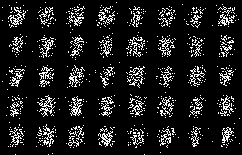}
\end{subfigure}
\begin{subfigure}{0.5\textwidth}
\centering
\includegraphics[scale=0.3]{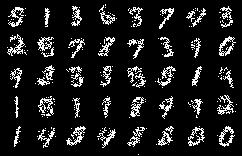}
\end{subfigure}\\
\begin{subfigure}{0.5\textwidth}
\centering
  \includegraphics[scale=0.3]{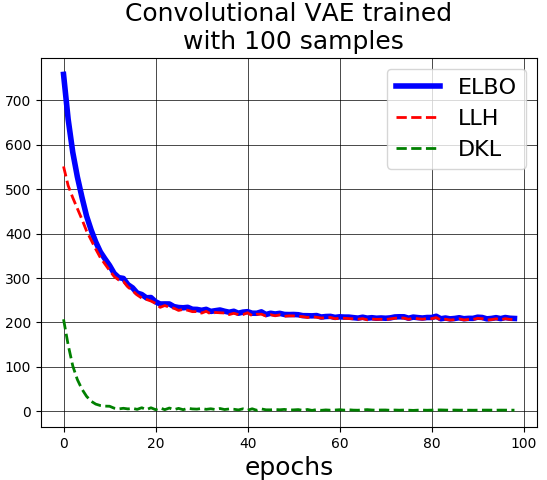}
\end{subfigure}
\begin{subfigure}{0.5\textwidth}
\centering
\includegraphics[scale=0.3]{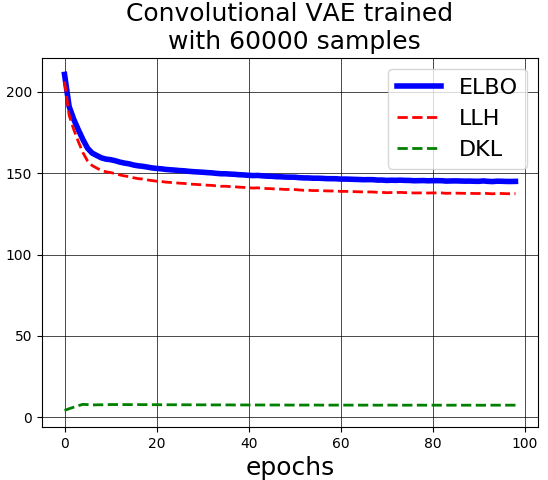}
\end{subfigure}
\caption{Samples obtained by decoding a sample from the prior distribution with two VAEs trained on 100 (top left) and 60000 (top right) samples from the MNIST database. Below we plot the reconstruction error (red dashed line) showing that although we are minimizing it, we cannot generate good quality images. Acronyms: $ELBO$ evidence lower bound; $D_{KL}$ Kullback-Lieber divergence and $LLH$ log-likelihood.}
\label{fig1}
\end{figure}

In this work we study how we can apply deep learning techniques when
the amount of data is very scarce. We simulate scenarios
where not only the amount of labeled data is scarce, but all the
available data. As mentioned before, some techniques can deal with
such scenarios. Bayesian modeling incorporates the uncertainty in the model
\cite{Bishop:2006:PRM:1162264}. However Bayesian neural networks are a
field under study and introduce several problems for which there is not
a wide well established solution: Monte Carlo integration, variational
approximations or sampling in high dimensional data spaces, among others. 
\newline

On the other hand, semi supervised learning techniques need a great amount of unlabeled
data to work well. For instance, the ladder network can achieve
impressive results with only 100 labeled samples in the MNIST task but
using 60000 unlabeled samples.
\newline

Finally, deep generative models (DGM) need great amounts of data to be
able of generate good quality samples. Figure \ref{fig1} shows a
Variational Auto-encoder (VAE) trained with 100 and 60000 samples. We
can see that although the reconstruction error is being minimized the
VAE with few samples is unable to generate good samples. 
\newline

To our knowledge, none of the above mentioned techniques (both semi supervised and DA
with DGM) has been applied disruptively to train neural networks
models in data scarce scenarios as the ones we propose. Moreover DA based on DGM has not
achieved impressive results in neural networks training with lots of
data. 
\newline

In this work we show that simple generative models as the
Restricted Boltzmann Machines (RBM)
\cite{Bengio:2009:LDA:1658423.1658424} clearly outperforms the ladder
network and DA based on a Deep Convolutional Variational
Auto-encoder. 

\section{Methodology}

In this work we simulate very scarce data scenarios. We train binary VAE and RBM using all the available samples. Details
on these models can be found at
\cite{Bengio:2009:LDA:1658423.1658424,1312.6114,1401.4082}. Once these
models are trained, we perform a sample generation following a MCMC procedure.

\subsection{Sample Generation}

For sample generation we rely on the theory of MCMC algorithms and define our transition operator as: 

\begin{equation}
\mathcal{T}(x'|x)=\int dh\, p(x'|h)\cdot p(h|x)
\end{equation}

Where $p(x'|h)$ and $p(h|x)$ represents the likelihood distribution of
an observed sample $x$ given a latent variable $h$ and, the posterior
distribution over the latent variable given an observed sample, respectively. We
will assume that this transition operator generates an ergodic Markov
Chain and thus as long as the number of generated samples goes to
infinity we will be sampling from the model distribution $p(x)$
\cite{neal1993probabilistic,Bishop:2006:PRM:1162264,DBLP:journals/corr/abs-1305-6663}. In case of VAEs, where the posterior distribution is approximated, see
\cite{1401.4082} appendix F for a proof of correctness.
\newline

In our models the likelihood distribution $p(x|h)$ is modeled with a
Bernoulli distribution. The posterior distribution is modeled with a
Bernoulli distribution for the RBM and with a factorized Gaussian
distribution for the VAE. For generating a sample we follow the
Contrastive Divergence \cite{carreira2005contrastive} algorithm which
is based on Gibbs Sampling but starting from an observed sample. As
example for generating 100 samples we follow algorithm \ref{algo1},
where $x$ is a sample from our dataset from which we will be
generating new samples and $N$ is the number of samples to
generate.\footnote{\scriptsize In case of VAE $p(h|x)$ is replaced by
  $q(h|x)$ which is the Variational Distribution. Note that although a
  Gibbs sampler depends on all the previous generated dimensions of a
  sample, in this case we can sample all the feature dimensions in
  parallel and thus our method is highly efficient.}

\begin{algorithm}[!t]
\SetAlgoLined
\textbf{Input}: $x$ , $N$\\
\textbf{Output}: generated\_chain\\
 generated\_chain=vector($N$)\\
 $x'=x$\\
 \For{$i=0$ until $N-1$ }{
  $h' \sim p(h|x')$\\
  $x' \sim p(x|h')$\\
  generated\_chain[$i$]$=x'$
  }
 \caption{Running MCMC algorithm for data generation}
\label{algo1}
\end{algorithm}

\begin{figure}[!b]
\begin{subfigure}{0.3\textwidth}
\centering
  \includegraphics[scale=0.25]{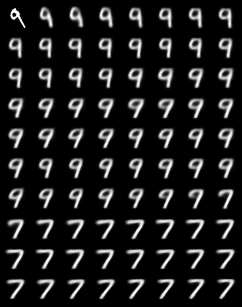}
\end{subfigure}
\begin{subfigure}{0.3\textwidth}
\centering
\includegraphics[scale=0.25]{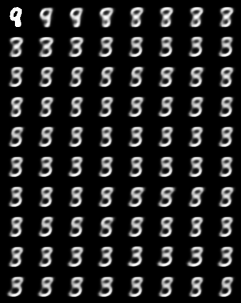}
\end{subfigure}
\begin{subfigure}{0.3\textwidth}
\centering
  \includegraphics[scale=0.25]{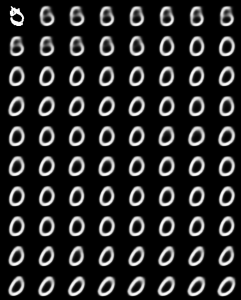}
\end{subfigure}
\caption{This figure shows $\mathbb{E}_{x\sim p(x|h)}\{x\}$ of a Markov Chain run with a VAE trained on 60000 samples, starting from a test image (image on the top left corner). We can clearly see how the generated samples can change from class as we run the algorithm. Image in the middle starts from an 8, then, the two next generated samples are 9s to after generate 8s and finally generate 3s. Sample on the right starts from a 5, generate some 5s and finally generate 0s.}
\label{fig2}
\end{figure}
\subsection{Labeling process}

We use the generated samples in two ways. As we stated, our approach
is based on training a classifier on a set of labeled samples using
additional generated samples from a VAE or a RBM. We associate the
generated samples with the same label as the sample from the data
distribution. In a first approach we use all the generated samples
(and denote this approach in the experiments with letter $n$). In the
second approach we classify the samples from the chain (using the same
classifier we are training) and only the correctly classified samples
are used for training (we denote this approach in the experiments with
letter $y$). This has a great impact, as shown in the experiments,
because long Markov Chains are likely to generate samples from other
classes, as shown in figure \ref{fig2}. 
\newline

\begin{figure}[!t]
\begin{subfigure}[c]{0.5
\textwidth}
\centering
  \includegraphics[scale=0.25]{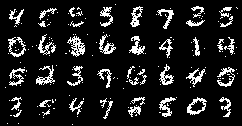}
\end{subfigure}
\begin{subfigure}[c]{0.5\textwidth}
\centering
  \includegraphics[scale=0.25]{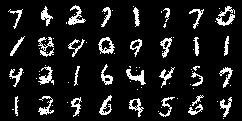}
\end{subfigure}
\caption{Samples from a bad (left) and good (right) RBM. Figure shows a sample from a MCMC chain of 1 step starting from a test sample.}
\label{fig3}
\end{figure}

Moreover, in case of the RBM we train two kind of models, named B-RBM
("bad RBM") and G-RBM ("good RBM"). The difference rely on the convergence
of the model, i.e., how is the quality of the generated samples, see
figure \ref{fig3}.  We expect that with a B-RBM the injected noise is
able to improve the generalization whereas the G-RBM is collapsing to
a part of the model space where no generalization improvement will be
obtained.  Basically we do not let the model achieve the same minimum for the case of the B-RBM as we do with the G-RBM.
\newline

\begin{figure}[!b]
\begin{subfigure}{\textwidth}
\centering
    \hspace*{\fill}
    {\includegraphics[scale=0.3]{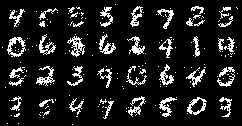}}
  \hfill
  {\includegraphics[scale=0.3]{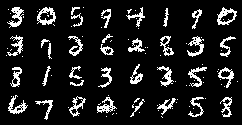}}\hspace*{\fill}
 \subcaption{B-RBM trained with 100 samples (left) and 1000 samples (right).} 
\end{subfigure}\newline
\begin{subfigure}{
\textwidth}
\centering
  \hspace*{\fill}
  {\includegraphics[scale=0.25]{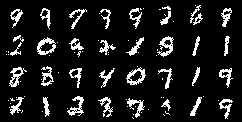}}
  \hfill
  {\includegraphics[scale=0.25]{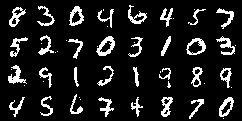}}
  \hspace*{\fill}
  \subcaption{ G-RBM trained with 100 samples (left) and 1000 samples (right).}
\end{subfigure}\newline

\begin{subfigure}{\textwidth}
\centering
\hspace*{\fill}
  {\includegraphics[scale=0.25]{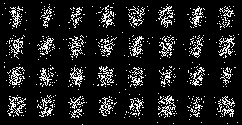}}
  \hfill
  {\includegraphics[scale=0.25]{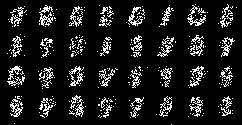}}
  \hfill
  {\includegraphics[scale=0.25]{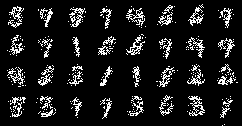}}
  \hfill
  {\includegraphics[scale=0.25]{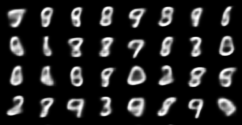}}
    \hspace*{\fill}
\subcaption{Samples from a VAE trained with 100 samples (left), 1000 samples (middle left), 10000 samples (middle right) and $\mathbb{E}_{x\sim p(x|h)}\{x\}$ on 10000 samples (right)}
\end{subfigure}\newline
\caption{Samples from our proposed generative models}
\label{fig4}
\end{figure}

Finally, figure \ref{fig4} shows images from the different trained
models in this work. We can clearly see how the VAE is able to
generate good quality samples only when more training samples are provided.

\section{Experiments}
For the experiments we use a binarized version from the MNIST
database. This database has 60000 training samples and 10000 test
samples. The pixels above $0.5$ are saturated to value $1.0$ and the
rest are saturated to $0.0$. In order to simulate a scarce data
scenario, we randomly select a small set of samples, and assume that only a
very small subset is labeled. We simulate three different scenarios
with a total of $100$, $1000$ and $10000$ samples where only $10$,
$100$ and $1000$ are labeled respectively. Note that for the first
scenario we have only $1$ labeled sample per class.
\newline

We use a binarized version of this database because, the expressions of the conditional distributions of the RBM models we use, are obtained assuming binary data distributions. Moreover, the VAE models for MNIST converge better when using Bernoulli decoders , ie binary cross entropy loss. \todo{aqi poner algo como que muchas tareas de texto usan datos binarizados, firma etc???}
\newline

We trained 3 models, two fully connected (FC) and one convolutional
(CNN). For fully connected we choose the following parameters, FC1:
784-1024-1024-10, FC2: 784-1000-500-250-250-250-10. For the
convolutional counterpart we use, CONV1:
32@3x3-64@3x3-128@3x3-512-512-10. In all the topologies
we inject Gaussian noise with $\sigma=0.3$ in the input and we use
batch norm (BN)\cite{DBLP:journals/corr/IoffeS15} and dropout
\cite{JMLR:v15:srivastava14a})
\newline

Tables \ref{tab3}, \ref{tab4} and \ref{tab5} show the error percentage
with the here proposed data augmentation showing that the B-RBM
clearly outperforms other approaches. We generate Markov chains of 500
and 1000 samples to increase the data set and train the
classifier\footnote{\scriptsize convolutional models on 10 labeled
  samples are trained with 850 instead of 1000 samples. Convolutional
  models for 100 and 1000 samples use chains of 100 samples. VAE model
  on 100 and 1000 samples for all the schemes generates 100 samples. We
  found a GPU-memory bottleneck because we performed a parameter
  update per batch with all its generated samples}. It is interesting to
see that although the deep FC (FC2) has worse performance than FC1
with 10 and 100 samples without DA, we can achieve better results in
case of 100 samples with FC2 when using our proposed method. \newline

We also see that a significant improvement is obtained with the most
scarce scenario (see table \ref{tab3}), where we are able to reduce
17\% error on CONV1 (check B-RBM option $y$ 1000 samples) and more
than 10\% in FC models (check B-RBM option $y$), which is the main
objective of this work.
\newline

\begin{table}[!b]\caption{Data Augmentation for 10 labeled samples}
\scriptsize
\centering
\begin{adjustbox}{max width=\textwidth}
\begin{tabular}
{c|c|cc|cc|cc|cc|cc|cc}
\hline
& \textbf{Baseline} &\multicolumn{4}{c}{\textbf{B-RBM}}                 & \multicolumn{4}{|c}{\textbf{G-RBM}}                 & \multicolumn{4}{|c}{\textbf{VAE}}                   \\
\textbf{Chain Length}& & \multicolumn{2}{c|}{500} & \multicolumn{2}{c|}{1000} & \multicolumn{2}{c|}{500} & \multicolumn{2}{c|}{1000} & \multicolumn{2}{c|}{500} & \multicolumn{2}{c}{1000} \\\hline
\textbf{Classify}  &   & y           & n         & y           & n          & y           & n         & y           & n          & y          & n          & y           & n          \\\hline
FC1           &    53.71    & 44.71       & 47.87     & \textbf{43.19}     & 47.56      & 53.49       & 52.55     & 53.91       & 52.5       & 80.25      & 51.49      & 84.48       & 50.56      \\
FC2         &    58.88      &\textbf{ 45.34 }      & 47.18     & 46.4        & 46.26      & 54.74       & 55.27     & 57.43       & 57.96      & 78.14      & 56.91      & 78.79       & 56.66      \\
CONV1     &      49.58      & 33.77       & 37.51     &\textbf{32.26}      & 36.69      & 41.260      & 38.94     & 40.12       & 40.66      & 39.35      & 41.56      & 44.86       & 41.47      \\     
\end{tabular} 
\end{adjustbox}
\label{tab3}
\end{table}
\begin{table}[!t]
\scriptsize
\centering
\caption{Data Augmentation for 100 labeled samples}
\begin{adjustbox}{max width=\textwidth}
\begin{tabular}{c|c|cc|cc|cc|cc|cc|cc}\hline
 & \textbf{Baseline} & \multicolumn{4}{c}{\textbf{B-RBM}}     & \multicolumn{4}{|c}{\textbf{G-RBM}}                           & \multicolumn{4}{|c}{\textbf{VAE}}                   \\
\textbf{Chain Length} & & \multicolumn{2}{c|}{500} & \multicolumn{2}{c|}{1000} & \multicolumn{2}{c|}{500} & \multicolumn{2}{c|}{1000}    & \multicolumn{2}{c|}{500} & \multicolumn{2}{c}{1000} \\\hline
\textbf{Classify}  &   & y          & n          & y                     & n      & y      & n          & y          & n           & y        & n         & y           & n          \\\hline
FC1       &    26.56        & \textbf{21.34}      & 21.61      & 21.41                 & 22.43                 & 26.83     & 26.51     & 28.41                & 26.86                & 52.01      & 37.00    &   -         & -          \\
FC2      &        28.39     & 19.72      & 21.31      & \textbf{18.66 }                & 22.31                 & 26.95     & 26.98     & 25.96                & 26.54                & 64.82      & 43.99    & -           & -          \\
CONV1      &     12.41      & 11.65      & 13.55      & \multicolumn{1}{c}{-} & \multicolumn{1}{c|}{-} & 11.36     & \textbf{11.25}     & \multicolumn{1}{c}{-} & \multicolumn{1}{c|}{-} & 58.35      & 30.14    & -           & -          \\        
\end{tabular}
\end{adjustbox}
\label{tab4}
\end{table}
\begin{table}[!t]
\scriptsize
\centering
\caption{Data Augmentation for 1000 labeled samples}
\begin{adjustbox}{max width=\textwidth}
\begin{tabular}
{c|c|cc|cc|cc|cc|cc|cc}
\hline
& \textbf{Baseline} &\multicolumn{4}{c|}{\textbf{B-RBM}}                                                                                    & \multicolumn{4}{c|}{\textbf{G-RBM}}                                                                      & \multicolumn{4}{c}{\textbf{VAE}}                   \\
\textbf{Chain Length} & &\multicolumn{2}{c|}{500}                                   & \multicolumn{2}{c|}{1000}                                  & \multicolumn{2}{c|}{500}                                   & \multicolumn{2}{c|}{1000}                    & \multicolumn{2}{c|}{500} & \multicolumn{2}{c}{1000} \\\hline
\textbf{Classify}  &   & y                           & n                           & y                           & n                           & y                           & n                           & y                    & n                    & y          & n          & y           & n          \\\hline
FC1         &    7.62      & \textbf{5.55} & 6.16 & 5.81 &6.04 &  5.86 &  5.91 &          5.79            &       5.97               & 24.13      & 12.42      &     -        &      -      \\
FC2         &    7.25      & 5.60 &  5.96 &  5.76 &  5.62 &  5.28 &  5.40 & \textbf{4.70}          & 5.49                & 36.19      & 18.00      &          -   &  -          \\
CONV1    &      3.11       &  3.26 & 3.89 & \multicolumn{1}{c}{-}        & \multicolumn{1}{c|}{-}        & \textbf{3.09}                        & 3.54                        & \multicolumn{1}{c}{-} & \multicolumn{1}{c|}{-} & 10.19      & 4.85       &       -      &   -         \\
\end{tabular}
\end{adjustbox}
\label{tab5}
\end{table}

Finally, Table \ref{tab6} shows a comparison with the ladder network. Ladder
network can be considered the state-of-the art on semi-supervised learning on this dataset\footnote{\scriptsize Recently other proposed methods have achieved better results, but they are based on GANs and we showed here that DGM are not suitable for these scenarios. For that reason we compare with ladder network.}. As can be seen we obtain better results on
the three scenarios.

\begin{table}[!t]
\scriptsize
\caption{A comparison with the ladder network. We represent error percentage.}
\centering
\begin{tabular}{c|ccc}
\hline
\textbf{Labeled Samples} & \textbf{10} & \textbf{100} & \textbf{1000} \\\hline
\textbf{Baseline}           & 58.88       & 28.39        & 7.25          \\
\textbf{Ladder Network}     & 48.85       & 24.74        & 6.96          \\
\textbf{RBM DA}             & \textbf{45.34 }      & \textbf{ 18.66}       &\textbf{ 5.60 }           
\end{tabular}
\label{tab6}
\end{table}
\todo{\textbf{Comentario revisor:} For the ladder network experiment, ladder network usually uses a lot of unlabeled data to improve the performance. Therefore, can authors provide some foresee that how the proposed method perform using more data generated data? Like provide the results on labeled samples 10000 in table 4.}
\todo{esto realmente no lo pusimos porque no funciona mejor. El tema de los modelos generativos es que no generan muestras que mejoren, al menos hasta lo que Roberto y yo Sabemos.}

\section{Conclusions}
We can draw several conclusions from this work. We first show that in
data scarcity scenarios simple generative models outperform deep
generative models (like VAEs). We also see that a B-RBM is
incorporating noise that is improving generalization. We can check
that G-RBM and VAE works better when we do not classify the generated
sample and this is in fact another way to incorporate noise into the
classifier. However B-RBM is the best of the three. This also means
that a generative model trained in this way (where latent variables
capture high detail) is unable to generate samples that improve
generalization. The G-RBM generates better quality images but is
unable to improve classification accuracy as the B-RBM does. \newline

This can also be noted when we add more training samples, where the
difference between the baseline and the here proposed DA is lower, as
with CNN. This is because the samples generated do not
incorporate additional information to the model and are either quite
similar between them or quite similar to the labeled samples. A possible hypothesis is that the generative model is collapsing to a part of the data feature space.
\newline

VAEs results were unexpected because despite the poor quality images
generated it can improve performance over the baseline. We got this
improvement always without classifying images, model $n$,  and only in the case
where few label samples are used. It is clear that the VAE is not a
good model for these scenarios. 
\newline

Finally, we also show that the here proposed approach outperforms and
is clearly an alternative to semi supervised learning in data scarcity
scenarios as shown in table \ref{tab6}. Another important advantage
is that RBM is robust and has a stable learning whether the ladder
network and GAN frameworks have several training challenges. The ladder
network has many hyper-parameters and its performance is really
sensible to little changes on them and the GANs are quite sensible to
hyper-parameters as well. 

\section{Acknowledgment}
We gratefully acknowledge the support of NVIDIA Corporation with the donation of two Titan Xp GPU used for this research. The work of Daniel Ramos has been supported by the Spanish Ministry of Education by project TEC2015-68172-C2-1-P. Juan Maroñas is supported by grant FPI-UPV.

\bibliographystyle{abbrv}

\begin{thebibliography}{10}

\bibitem{Bengio:2009:LDA:1658423.1658424}
Y.~Bengio.
\newblock Learning deep architectures for ai.
\newblock {\em Found. Trends Mach. Learn.}, 2(1):1--127, Jan. 2009.

\bibitem{DBLP:journals/corr/abs-1305-6663}
Y.~Bengio et~al.
\newblock Generalized denoising auto-encoders as generative models.
\newblock In {\em Advances in Neural Information Processing Systems 26}, pages
  899--907. Curran Associates, Inc., 2013.

\bibitem{Bishop:2006:PRM:1162264}
C.~M. Bishop.
\newblock {\em Pattern Recognition and Machine Learning}.
\newblock Springer-Verlag, 2006.

\bibitem{carreira2005contrastive}
Carreira-Perpinan et~al.
\newblock On contrastive divergence learning.
\newblock In {\em Aistats}, volume~10, pages 33--40. Citeseer, 2005.

\bibitem{1506.02158}
Y.~Gal and Z.~Ghahramani.
\newblock Bayesian convolutional neural networks with {B}ernoulli approximate
  variational inference.
\newblock In {\em 4th International Conference on Learning Representations
  (ICLR) workshop track}, 2016.

\bibitem{1406.2661}
I.~Goodfellow et~al.
\newblock Generative adversarial nets.
\newblock In {\em Advances in Neural Information Processing Systems 27}, pages
  2672--2680. Curran Associates, Inc., 2014.

\bibitem{hinton16speechprocessing}
G.~Hinton et~al.
\newblock Deep neural networks for acoustic modeling in speech recognition.
\newblock {\em IEEE Signal Processing Magazine}, 29:82--97, 2012.

\bibitem{DBLP:journals/corr/IoffeS15}
S.~Ioffe et~al.
\newblock Batch normalization: Accelerating deep network training by reducing
  internal covariate shift.
\newblock In {\em Proceedings of the 32Nd International Conference on
  International Conference on Machine Learning - Volume 37}, ICML'15, pages
  448--456. JMLR.org, 2015.

\bibitem{1312.6114}
D.~P. Kingma et~al.
\newblock Auto-encoding variational bayes, 2013.

\bibitem{DBLP:journals/corr/abs-1301-3781}
T.~Mikolov et~al.
\newblock Efficient estimation of word representations in vector space.
\newblock 2013.

\bibitem{neal1993probabilistic}
R.~M. Neal.
\newblock Probabilistic inference using markov chain monte carlo methods.
\newblock 1993.

\bibitem{NIPS2015_5947}
A.~Rasmus et~al.
\newblock Semi-supervised learning with ladder networks.
\newblock In {\em Advances in Neural Information Processing Systems 28}, pages
  3546--3554. Curran Associates, Inc., 2015.

\bibitem{1506.02640}
J.~Redmon et~al.
\newblock You only look once: Unified, real-time object detection.
\newblock In {\em 2016 {IEEE} Conference on Computer Vision and Pattern
  Recognition, {CVPR} 2016, Las Vegas, NV, USA, June 27-30, 2016}, pages
  779--788, 2016.

\bibitem{1401.4082}
D.~J. Rezende et~al.
\newblock Stochastic backpropagation and approximate inference in deep
  generative models.
\newblock In {\em Proceedings of the 31st International Conference on
  International Conference on Machine Learning - Volume 32}, ICML'14, pages
  II--1278--II--1286. JMLR.org, 2014.

\bibitem{JMLR:v15:srivastava14a}
N.~Srivastava et~al.
\newblock Dropout: A simple way to prevent neural networks from overfitting.
\newblock {\em Journal of Machine Learning Research}, 15:1929--1958, 2014.

\bibitem{DBLP:journals/corr/SutskeverVL14}
I.~Sutskever et~al.
\newblock Sequence to sequence learning with neural networks.
\newblock In {\em Advances in Neural Information Processing Systems 27}, pages
  3104--3112. Curran Associates, Inc., 2014.

\bibitem{DBLP:journals/corr/SzegedyIV16}
C.~Szegedy et~al.
\newblock Inception-v4, inception-resnet and the impact of residual connections
  on learning.
\newblock 2016.

\bibitem{NIPS2017_6872}
Tran et~al.
\newblock A bayesian data augmentation approach for learning deep models.
\newblock In I.~Guyon et~al., editors, {\em Advances in Neural Information
  Processing Systems 30}, pages 2797--2806. Curran Associates, Inc., 2017.

\bibitem{1606.05328}
A.~van~den Oord et~al.
\newblock Conditional image generation with pixelcnn decoders.
\newblock In {\em Advances in Neural Information Processing Systems 29}, pages
  4790--4798. Curran Associates, Inc., 2016.

\end{thebibliography}
\FloatBarrier

\end{document}